\documentclass[letterpaper, 10 pt, conference]{ieeeconf}  

\IEEEoverridecommandlockouts                              
\usepackage{cite}
\usepackage{amsmath,amssymb,amsfonts}
\usepackage{algorithmic}
\usepackage{graphicx}
\usepackage{textcomp}
\usepackage{xcolor}
\def\BibTeX{{\rm B\kern-.05em{\sc i\kern-.025em b}\kern-.08em
    T\kern-.1667em\lower.7ex\hbox{E}\kern-.125emX}}

\usepackage{hyperref}
\hypersetup{
  pdftitle={Transferring Tactile Data Across Sensor Modalities},
  pdfauthor={Wadhah Zai El Amri, Malte Kuhlmann, Nicolás Navarro-Guerrero},
  pdfsubject={Robotics (cs.RO); Artificial Intelligence (cs.AI),},%
  pdfkeywords={},%
  colorlinks=true,
  linktoc=all,
  linkcolor=darkgray,
  citecolor=darkgray,
  urlcolor=black,
}
\usepackage[protrusion=true,expansion=true]{microtype}
\usepackage{textcomp}
\usepackage{multirow}
\usepackage{blindtext}
\usepackage{footmisc}
\usepackage{xcolor}
\usepackage{todonotes}

\begin{document}

\title{\LARGE \bf
Transferring Tactile Data Across Sensors}

\author{Wadhah Zai El Amri$^{1}$, Malte Kuhlmann$^{1}$ and Nicolás Navarro-Guerrero$^{1}$
\thanks{$^{1}$L3S Research Center, Hanover, Germany
        {\tt\small \{wadhah.zai, malte.kuhlmann, nicolas.navarro\}@l3s.de }}%
}

\maketitle

\begin{abstract}
Tactile perception is essential for human interaction with the environment and is becoming increasingly crucial in robotics. Tactile sensors like the BioTac mimic human fingertips and provide detailed interaction data. Despite its utility in applications like slip detection and object identification, this sensor is now deprecated, making many existing datasets obsolete. This article introduces a novel method for translating data between tactile sensors by exploiting sensor deformation information rather than output signals. We demonstrate the approach by translating BioTac signals into the DIGIT sensor. Our framework consists of three Steps: first, converting signal data into corresponding 3D deformation meshes; second, translating these 3D deformation meshes from one sensor to another; and third, generating output images using the converted meshes. Our approach enables the continued use of valuable datasets.
\end{abstract}


\section{Introduction}
Tactile feedback is gaining significant attention in robotics\cite{Dahiya2010TactileRobots, Navarro-Guerrero2023VisuoHaptic}. 
Tactile sensors leverage various information modalities, come in diverse shapes and sizes, and are implemented in a wide range of technologies. This diversity makes the exchange of data and trained models challenging. Moreover, as sensor technology improves, datasets become deprecated. 
For instance, BioTac by SynTouch, was a high-end tactile sensor, designed like a human fingertip. It has an elastomer covering a rigid core filled with an incompressible conductive fluid. The sensor outputs voltage readings from 19 internal electrodes, capturing changes in the fluid.  These readings are processed as time-series signal data~\cite{Wettels2007,Wettels2014,ZaiElAmri2024Optimizing}. The BioTac has been proven useful in various applications such as detecting object slips and the direction of slips~\cite{Garcia2019BioTacDataset, Zapata2019BioTacDataset} or identifying objects~\cite{Xu2013}. However, this sensor is now deprecated. Consequently, many existing datasets, such as the BioTac SP direction of slip dataset ~\cite{Zapata2019BioTacDataset}, the BioTac SP grasp stability dataset \cite{Garcia2019BioTacDataset} or BioTac 2P grasp stability dataset \cite{chebotar2016bigs}, are now obsolete.
Hence, there is a need to convert existing datasets into formats compatible with newer sensor modalities, allowing researchers to leverage intrinsic information still relevant to specific tasks.


Although some research attempts to transfer between modalities, for instance, Lee et al.~\cite{lee2019touching} developed a framework to generate tactile images from the GelSight sensor using digital camera images of various cloth materials and vice versa, or the ViTac dataset~\cite{luo2018vitac}, used to train the networks, includes labeled images from the GelSight sensor and a digital camera of 100 fabric pieces. Unfortunately, these approaches require labeled data from both sensor types, and the modality of the source and target sensors is the same, i.e., vision.

In contrast, our approach focuses on transferring the encoded information and knowledge at the deformation level, eliminating the need for congruence in the tactile output of the sensors. To our knowledge, this is the first article showing transfer between two sensors with entirely different sizes, shapes, and output spaces. Additionally, our approach does not require end-to-end labeled sensor outputs and can be generalized to any force or orientation. Therefore, our method enables the utilization of any existing datasets. 
We demonstrate our approach by transferring low-resolution tactile (time series) data from a BioTac into a vision-based DIGIT sensor~\cite{Lambeta2020DIGIT}. The DIGIT sensor is a low-cost, high-resolution, vision-based tactile sensor.

\section{Methodology}\label{sec:methods}
Transferring between different modalities poses challenges due to data representations and encoding variations. To address these issues, we propose a three-step solution, depicted in Fig.~\ref{fig:full_pipeline_networks}. Step I: We initially predict the BioTac surface deformation from the BioTac input signals. Step II: We convert the BioTac surface mesh deformation to DIGIT surface mesh deformation since the physical interaction of both sensors can be modeled by a mesh deformation independently of sensor output modality. Step III: We generate the DIGIT sensor image from the converted deformation. 

\subsection*{Step I: Predicting BioTac Mesh Deformation}
We adopt a similar methodology to that proposed by Narang et al.~\cite{narang2021isaac}. We train a disentangled variational autoencoder ($\beta$-VAE)~\cite{Higgins2016betaVAELB} to reconstruct the BioTac sensor outputs. This network is denoted as Signal VAE BioTac ({\textit{SVB}}).

We also train another $\beta$-VAE to reconstruct the 3D mesh deformation of the BioTac sensor, denoted as Mesh VAE BioTac ({\textit{MVB}}). To model these deformations and collect the dataset used for this task, we employ the Isaac Gym BioTac model~\cite{narang2021isaac}. 

Next, we train an MLP network to map between the latent vectors of the \textit{SVB} and the \textit{MVB} network, referred to as Signal to Mesh Projection Network ({\textit{S2MPN}}). For this latent space mapping, we use the publicly available dataset collected by Narang et al.~\cite{narang2021isaac}. 

\subsection*{Step II: Modeling of Mesh Deformation}
In this Step, we train a third $\beta$-VAE~\cite{Higgins2016betaVAELB}, with the same architecture used for the \textit{MVB} network, this time to reconstruct the DIGIT 3D deformations. This network is denoted as Mesh VAE DIGIT ({\textit{MVD}). Afterward, we train an MLP network to map the latent space of the already trained \textit{MVB} encoder network in Step I to the latent space of the trained \textit{MVD} encoder network, we denote this network as Mesh to Mesh Projection Network (\textit{M2MPN}). 

To train the \textit{M2MPN}, we collected over 50K unique mesh deformations for both BioTac and DIGIT using the Isaac Gym simulator. To ensure alignment between the two mesh datasets, we maintained consistent force, angle, and position parameters for each interaction pair since labeled and matching BioTac DIGIT meshes are required for training the \textit{M2MPN}. To address the difficulty in representing side touches arising from the differing shapes of the DIGIT and BioTac sensors, we rotate and translate the BioTac sensor on its axis within its horizontal plane. This transformation mimics the unfolding of the BioTac elastomer to align with and cover the flat surface of the DIGIT sensor. This solution ensures that forces and deformations resulting from side touches correspond with the other sensor.

\subsection*{Step III: From Surface Deformation to DIGIT's output}
The DIGIT sensor contains a camera that captures light reflections caused by surface deformations on the silicon-based gel pad when interacting with other objects. We adapt the simulation model \textit{Taxim}~\cite{si2022taxim} to simulate DIGIT images. Taxim calculates a height map of the gel pad using object point clouds from which the corresponding DIGIT image is estimated. We extend this approach to determine the height map using the deformation mesh.

\begin{figure}[htbp]
  \centering
  \includegraphics[width=0.6\columnwidth]{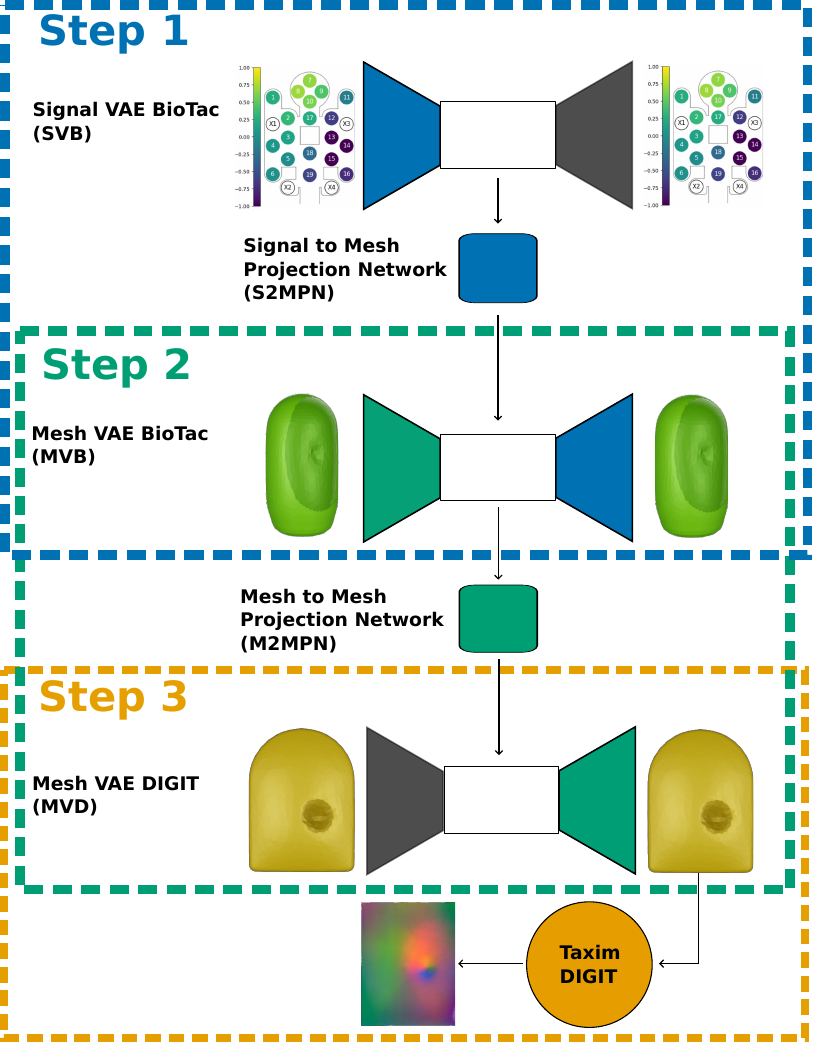}
  \caption{Framework for translating BioTac signals into DIGIT images. Step 1: Convert BioTac input signals to BioTac surface deformation. Step 2: Convert BioTac surface mesh deformation to DIGIT surface mesh deformation. Step 3: Generate DIGIT's output from the surface mesh deformation.}
  \label{fig:full_pipeline_networks}
\end{figure}

\section{Results and Conclusion}
As a proof of concept, we trained the networks with our paired generated dataset collected with both sensors while interacting with a spherical indenter. We then selected a test set from available real BioTac signal data~\cite{narang2021isaac} and converted it to DIGIT output images. Fig.~\ref{fig:results_samples} shows qualitative results of five selected BioTac signals that we converted to DIGIT images using our framework.

\begin{figure}[htbp]
  \centering
  \includegraphics[width=0.95\columnwidth]{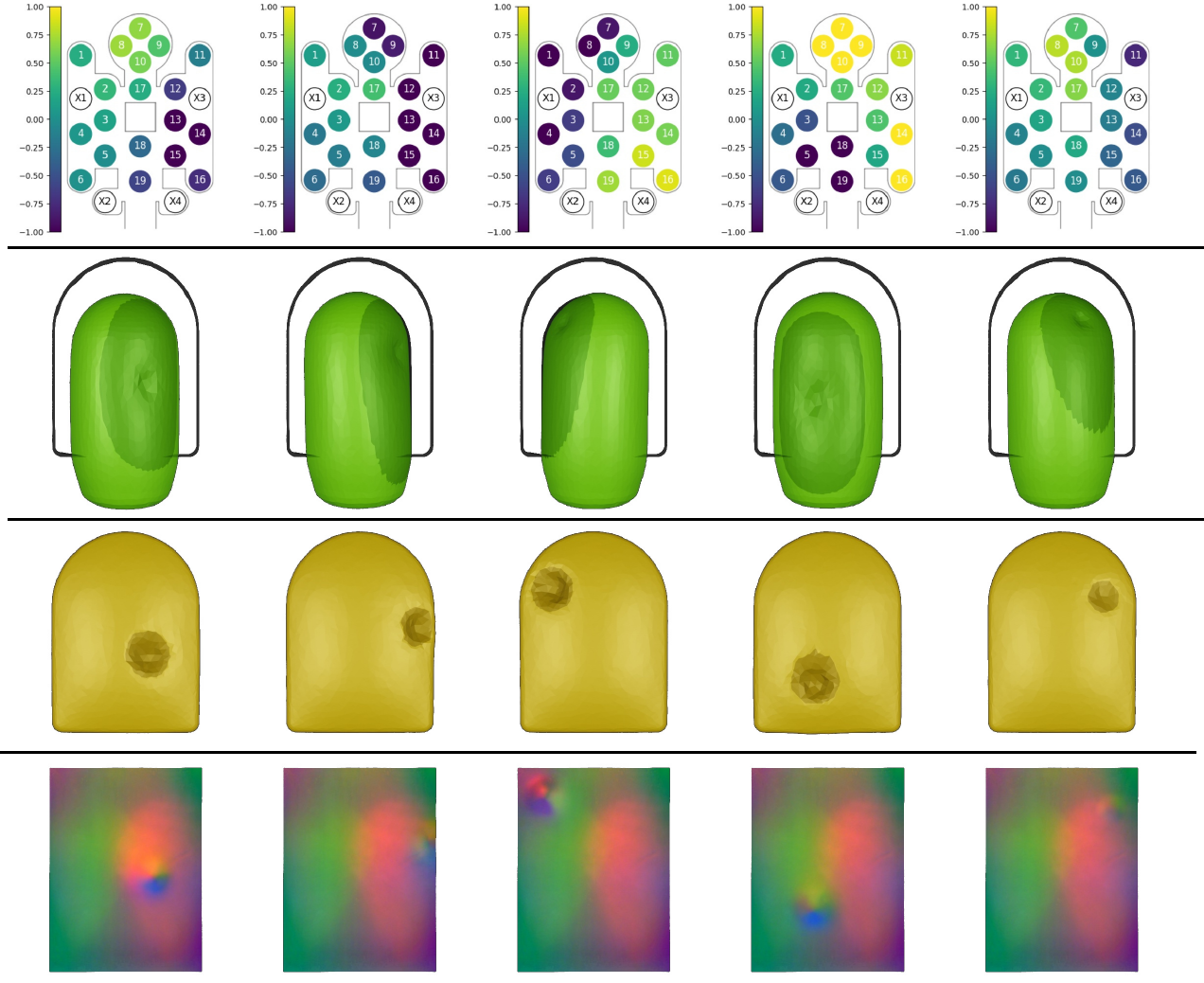}
  \caption{Converted samples. First row: Real electrode values. Second row: Ground-truth BioTac mesh deformations. The outer frame represents the ``unfolded'' BioTac surface. Third row: Converted DIGIT mesh deformations. Fourth row: DIGIT output images. The third and fourth rows were generated using the first row as input.}
  \label{fig:results_samples}
\end{figure}

To conclude, we address the problem of translating tactile sensor outputs regardless of shape, size, and output representation mismatch, which we demonstrate on the BioTac and the DIGIT sensors. We are preparing an extended version of this work for ICRA2025, including other indenters and a more detailed explanation of the approach. In future work, we plan to refine the approach and convert popular BioTac datasets to DIGIT images.


\addtolength{\textheight}{-195mm}   

\bibliographystyle{IEEEtran}
\bibliography{references.bib}

\end{document}